# Ambiguity is the last thing you need

**Emily Chivers**

emilychivers30@gmail.com

**Shawn Curran**

shawn.curran@jylo.ai

## Abstract

Clear legal language forms the backbone of a contract for numerous reasons. Disputes often arise between contract parties where ambiguous language has been used and parties often disagree on the meaning or effect of the words. Unambiguous language can also be important where there is an imbalance of bargaining strength between the parties, for instance where a business is contracting with a consumer, where the law actually requires plain language to be used. Thus, plain language minimises misinterpretation and [1] prevents future litigation. Contracts become ambiguous when the language used is vague, imprecise, or open to multiple interpretations and this is due to the vast number of synonyms in the English Language which creates differences in interpretation between the meaning of the language. Ambiguity has always formed a prevalent issue in case-law, with a large percentage of cases based on ambiguous language. Thus, from an outside perspective the legal sector should look forward to ways of reducing this.

## 1 Thesis Proposition

### 1.1 Introduction

There are various future tools which one utilise to eradicate the ambiguity prevalent within case-law judgements. One such tool is using programming languages instead of regular English syntax in which contracts are currently written. This would subsequently remove ambiguity in legal contracts due to the reduced syntax of a declarative programming language. Simultaneously this approach, in principle, would also mitigate unnecessary cases rooted in misunderstandings between two parties and adopting such a method could greatly simplify the legal contracting process. The structured syntax of programming languages tends well towards a use in legal contracts, leaving no room for alternate phrasing and thus no room for ambiguity. For example, the following common legal contractual phrase:

"This agreement constitutes the entire understanding between the parties and supersedes all prior negotiations, representations, or agreements."

Could be converted into the following C# code:

```
using System;
using System.Collections.Generic;

public class Agreement
{
    private List<string> priorAgreements;
    private string currentAgreement;

    public Agreement(string agreementText)
    {
        priorAgreements = new List<string>();
        currentAgreement = agreementText;
    }

    public void AddPriorAgreement(string agreementText)
    {
        priorAgreements.Add(agreementText);
    }

    public bool IsCurrentAgreementSuperseding()
    {
        if (priorAgreements.Count > 0)
        {
            foreach (string prior in priorAgreements)
            {
                Console.WriteLine($"Prior agreement superseded: {prior}");
            }
            priorAgreements.Clear(); // All prior agreements are now considered superseded
            return true;
        }
        return false; // No prior agreements to supersede
    }

    public void DisplayCurrentAgreement()
    {
        Console.WriteLine($"This Agreement constitutes the entire understanding: {currentAgreement}");
    }
}

class Program
{
    static void Main()
    {
        Agreement agreement = new Agreement("This is the current understanding between the parties.");

        // Adding prior agreements to simulate previous negotiations or agreements
```

```
        agreement.AddPriorAgreement("Previous agreement on terms.");
        agreement.AddPriorAgreement("Representation made prior to the final agreement.");

        if (agreement.IsCurrentAgreementSuperseding())
        {
            Console.WriteLine("The current agreement supersedes all prior agreements.");
        }

        agreement.DisplayCurrentAgreement();
    }
}
```

This simplification from plain language to a programming language has numerous advantages which are listed in the section below.

## 1.2 Advantages

The use of programming languages for legal contracts has a multitude of advantages in eradicating ambiguity. This encompasses and is not limited to:

1. The defined conditions that programming languages, such as C#, operate on ensure that the contract consists of clear conditions and outcomes, leaving no room for ambiguity. This would, in principle, significantly reduce the complexity of legal contracts today as the principles of the contract would be clearly defined, leading to reduced ambivalence.

2. The exact syntax and specification of programming languages leaves no room for alternative phrasing which, for plain language contracts, creates disparities between the views exhibited from the two parties. Through this programming languages introduce a standardised language which uniformly defines and applies terms. This differs from natural language and ensures both parties interpret the terms identically.

3. Programming languages also enable contracts to be executed when predefined conditions are met. This reduced potential disputes over interpretation, ensuring that obligations are fulfilled without ambiguity. Additionally, executability increases efficiency and decreases the large volume of time spent reading lengthy plain language contracts.

4. The use of programming languages in contracts also enhances the verification process and error detection as code can be run, unlike a plain language contract, thus reducing the likelihood of disputes arising from conflicting terms.

These points all evidently highlight how advantageous the switch to the use of programming languages would be in the legal sector. Namely, the contrast between the ambiguous nature of English syntax and the defined syntax of a programming language would largely disperse the complex and ambiguous legal language prevalent in many contracts.

## 1.3 Disadvantages and potential solutions

Though in principle the switch to the use of programming language from plain language for the purpose of unambiguous contracts may seem purely beneficial at first glance, there are disadvantages with the implementation process that would arise if this theory was put into practice. This includes:

1. The difficulty of educating a new generation of lawyers to be fluent and competent in a programming language. Writing a contract in a programming language requires greater knowledge than writing a contract in plain language, which increases complexity and can limit adoption to those without specialist knowledge.

2. The discrepancies between the syntax of different programming languages.

- A potential solution would be establishing a countrywide or worldwide universal programming languages that all lawyers wrote their contracts in. Contrastingly, it would be difficult to decide on what exact language to use as programming languages evolve rapidly and if a language is decided on, it may not always be the best language to use in the future.

3. The reduced syntax can be read as a disadvantage, as though natural language contracts have more room for misinterpretation, code leaves little room for flexibility. To further, this rigidity might create issues in contract interpretation as many cases often rely on the wide interpretation of legal language, though this is a flaw in the system.

4. One must also consider the varying laws in different countries that lie under the contract, meaning that a contract in one country could have a different meaning or effect in another country due to said laws.

- A potential solution would be to have a block of code which essentially represents the laws of the country, and this can be executed in parallel with the programmed contract, acting in the same way as a plain language legal contract.

Similar to the advantages of using programming languages in contracts, this theory also contains some disadvantages which would need to be carefully considered if such a theory was to be implemented in practice, namely the decision on a universal programming language.

# 2 Programming Languages

## 2.1 The maximalist nature of modern programming languages

The [2] minimalism that can be achieved in programming language design tends it towards a use in legal contracts. A significant concern for this argument is the trend towards maximalism in programming languages. Programming languages are designed for simplicity and efficiency, aiming to minimize redundant code. Despite this, many programming languages are adopting a maximalist approach where they introduce new features as they become threatened by the competitors around. This creates unnecessary tools which are intended to make programming accessible for all by reducing complexity. Instead, from a legal point of view, this would push the programming languages closer towards recognisable English syntax. This growth in programming language syntax and methods could result in such languages becoming so close to English that their use in contracts becomes irrelevant. Therefore, striking a balance between innovation and simplicity would be crucial to maintain the effectiveness of programming languages in legal contracts.

## 2.2 Executability of Programming contracts

As highlighted in Section 1.1, a major advantage of writing contracts in programming code is the ability to execute the program to determine the output without the ambiguity of reading into the language in the contract. This ability would reduce the amount of time lawyers spend reading contracts, and the alternative programming approach would involve pre-setting variables then executing the code to observe the output. These executable programming contracts, or smart contracts, would [3] "offer the promise of seamless execution without human intervention." Though the executability of contracts would be advantageous, [4] "even minor flaws or oversights [in the code] could lead to severe consequences such as unauthorised access [being granted to the wrong individual]."

## 2.3 High level vs Low level languages for this purpose

In principle, if the writing of contracts was to be switched from plain language to programming languages, then certain factors regarding the programming language must be discussed, this encompasses whether a high-level or low-level programming language should be used. This decision requires weighting of a variety of factors:

1. **Understandability** – as lawyers most likely will not predispose deep technical knowledge of programming languages, the most blatant choice for this category would be the use of a high-level language as it is simpler to understand and has increased syntax, making it closer to the English language than a low-level language. However, arguably, the point of this paper is to put forward a point of view that legal contracts should move away from the complex nature of the language, and towards the reduced syntax of programming languages. Thus, one could argue that the low-level languages could be more useful for this purpose as it has reduced syntax, leaving less room for ambiguity.
2. **Abstraction** – high-level programming languages would be an obvious choice for this as they provide higher levels of abstraction, making the code easier to understand (as explained above).
3. **Features** – high-level programming languages would again be the obvious choice here as many have complex structures, leading to higher capabilities, therefore making it easier to apply to a legal context.
4. **Accessibility** – as there are fewer professionals trained in low level program coding; it would be harder to train a generation of lawyers in low-level languages compared to high level languages. Thus, the logical choice again would be high-level languages as more professionals in the industry are fluent in these languages, leading to an easier implementation of programming languages in contracts.

In principle, high-level programming languages appear as the most logical language for coding legal contracts due to its high level of abstraction, increased understandability and accessibility compared to low level languages. Contrastingly, it is important to consider that low-level languages are the furthest away from the English language, and thus in theory would be better for the purpose of reducing ambiguity by reducing the volume of syntax in a contract. Ultimately, the reduced syntax results in lower capabilities of low-level languages, therefore a high-level language would be the most suitable choice, as it provided a balance of understandability and simplicity.

# 3 Research

## 3.1 Method of Conducting Research

All of the following research was carried out in the model agnostic platform - www.jylo.ai.

As an initial note, the research of the cases was conducted in the way detailed below to both test the capabilities of LLMs (GPT4), and to reduce the time spent searching through large judgements documents.

The research detailed below was conducted by initially using Llama3.1 and general research to collate a list of cases where ambiguous language was used, (leading to the two parties of the civil cases arguing over the interpretation of such phrasing). Once the cases had been selected, a GPT4 LLM model was used to search through the judgement, and was given the following prompt:

"Firstly, summarise this entire case. Then extract from this judgement three fields. The first is the extract of the language from the contract that is at issue in this case. The second is that language converted to deterministic code using the language C#. The third is a summary of the reason the language is ambiguous."

Jylo [5] then outputs the results in a format that can be seen below.

1. **Extract of the Contract Language at Issue:**

"The Rent payable under this Lease will be reviewed in accordance with this paragraph 3 on each of the Review Dates and such Rent payable from and including each such Review Date shall be the Revised Rent which shall be calculated as follows:

Revised Rent = Rent payable prior to the Review Date (disregarding any suspension of Rent) x Revised Index Figure / Base Index Figure"

2. **Contract Language Converted to Deterministic C# Code:**

Here's a C# representation of the formula:

```csharp
public class RentCalculator
{
   public double CalculateRevisedRent(double previousRent, double revisedIndexFigure, double baseIndexFigure)
   {
      if (baseIndexFigure == 0)
      {
         throw new DivideByZeroException("Base index figure must not be zero.");
      }

      return previousRent * (revisedIndexFigure / baseIndexFigure);
   }
}
```

3. **Summary of the Reason for Ambiguity:**

The ambiguity in the contract language arises from the exponential effect the Formula has on rent increases. When applied literally, the Formula results in rent escalating at a rate significantly higher than the increase in the RPI (Retail Prices Index), leading to increases that are described as irrational and arbitrary. The ambiguity stems from whether this outcome was intended or a result of a drafting error. This ambiguity is addressed under the "Chartbrook principle," which allows for correcting clear errors if both the error and the intended meaning are evident. In this case, it was concluded that the intended rent increases were to align proportionately with RPI changes over time, which the original language failed to achieve.

After this has been outputted, the search tool (a simple ctrl+F) is then used to search the judgement for the ambiguous language referenced by the AI to ensure no hallucinations have occurred, which as stated in this [6] paper, can cause issues within the legal industry, due to the use of incorrect information provided by the LLMs. Below is an example of the search carried out on a judgement.

The example below is the case of MonSolar IQ Ltd v Woden Park Ltd, and as produced by the LLM above, the ambiguous language extract has been quoted as the following:

"The Rent payable under this Lease will be reviewed in accordance with this paragraph 3 on each of the Review Dates and such Rent payable from and including each such Review Date shall be the Revised Rent which shall be calculated as follows:

Revised Rent = Rent payable prior to the Review Date (disregarding any suspension of Rent) x Revised Index Figure / Base Index Figure"

As stated above, the ctrl+F feature is then used and the [7] judgement is then searched for the phrase output from the LLM and the below phrase is found which matches the quoted produced from the LLM, showing no hallucinations have occurred.

14. Paragraph 3 provides for the rent to be reviewed in accordance with the Formula in

"**Review of Rent**

The Rent payable under this Lease will be reviewed in accordance with this par

Revised Rent = Rent payable prior to the Review Date (disregarding any suspe
Base Index Figure"

15. By reading in the relevant defined terms, the Formula can be more simply expresse

"Revised Rent = Previous year's Rent x May RPI for current year
May 2013 RPI"

## 3.2 Comparison between LLMs

Due to the nature of this paper, the use of AI was high, and thus the use of models for each purpose was carefully considered to ensure no hallucinations occurred. The primary LLM used for assistance in this paper was the platform [8] Jylo which includes GPT4. This was an obvious choice as much of the research of cases involved searching through large judgement documents and extracting quotes of ambiguity from these documents. For this purpose, it was vital to use a highly accurate model, such as GPT4 to reduce the risk of hallucinations, along with the Jylo human-in-the-loop verification process to further mitigate the hallucination risk. For lower accuracy and general questions, models such as Llama 3.1 were used as information provided from these questions was less important and was checked and backed up by alternate sources. The data below shows the rankings of the LLMs based off a variety of features, and clearly showcases the capabilities of GPT4:

| | FEATURES | | QUALITY | PRICE | OUTPUT TOKENS/S | LATENCY | |
|---|---|---|---|---|---|---|---|
| MODEL | CREATOR | CONTEXT WINDOW | INDEX Normalized avg | BLENDED USD/1M Tokens | MEDIAN Tokens/s | MEDIAN First Chunk (s) | FURTHER ANALYSIS |
| o1-preview | OpenAI | 128k | 85 | $26.25 | 31.4 | 32.36 | Model Providers |
| o1-mini | OpenAI | 128k | 82 | $5.25 | 65.8 | 15.32 | Model Providers |
| GPT-4o | OpenAI | 128k | 77 | $4.38 | 96.1 | 0.46 | Model Providers |
| GPT-4o (May '24) | OpenAI | 128k | 77 | $7.50 | 101.8 | 0.46 | Model Providers |
| GPT-4o mini | OpenAI | 128k | 71 | $0.26 | 85.8 | 0.50 | Model Providers |
| Llama 3.1 405B | Meta | 128k | 72 | $4.50 | 24.1 | 0.81 | Model Providers |
| Llama 3.2 90B (Vision) | Meta | 128k | 66 | $0.90 | 40.5 | 0.50 | Model Providers |
| Llama 3.1 70B | Meta | 128k | 65 | $0.88 | 65.0 | 0.48 | Model Providers |
| Llama 3.2 11B (Vision) | Meta | 128k | 54 | $0.19 | 120.2 | 0.31 | Model Providers |
| Llama 3.1 8B | Meta | 128k | 53 | $0.15 | 167.4 | 0.38 | Model Providers |
| Llama 3.2 3B | Meta | 128k | 47 | $0.08 | 158.2 | 0.36 | Model Providers |
| Llama 3.2 1B | Meta | 128k | 27 | $0.05 | 556.3 | 0.35 | Model Providers |
| Gemini 1.5 Pro (Sep) | Google | 2m | 80 | $2.19 | 60.2 | 0.80 | Model Providers |
| Gemini 1.5 Flash (Sep) | Google | 1m | 73 | $0.13 | 208.5 | 0.39 | Model Providers |
| Gemma 2 27B | Google | 8k | 61 | $0.80 | 54.5 | 0.50 | Model Providers |
| Gemma 2 9B | Google | 8k | 46 | $0.20 | 118.2 | 0.31 | Model Providers |
| Gemini 1.5 Flash (May) | Google | 1m | | $0.13 | 310.0 | 0.34 | Model Providers |
| Gemini 1.5 Pro (May) | Google | 2m | | $5.25 | 65.8 | 0.84 | Model Providers |
| Gemini 1.5 Flash-8B | Google | 1m | | $0.07 | 285.1 | 0.37 | Model Providers |
| Claude 3.5 Sonnet (June) | ANTHROP\C | 200k | 77 | $6.00 | 54.9 | 0.91 | Model Providers |
| Claude 3 Opus | ANTHROP\C | 200k | 70 | $30.00 | 26.2 | 1.90 | Model Providers |
| Claude 3 Haiku | ANTHROP\C | 200k | 54 | $0.50 | 123.8 | 0.54 | Model Providers |

The full table can be viewed from this link: [9] LLM ranking.

### 3.3 Purpose of research

The purpose of the conversion of ambiguous language to C# is to both support the thesis on the transfer from plain English to programming languages in the legal context, but also to illustrate how some ambiguous phrasing, when converted to C# code, simply cannot run as ambiguity through absence cannot be converted into C# code. In principle the flow of a contract and its language should be precise and logical, therefore should be able to be implemented into a logical program flow. Thus, if the ambiguous excerpt cannot be converted to C# code, then it simply lacked clarity in the first instance and should not have been implemented. The table below shows the research that was conducted into certain cases where language lacked clarity, and thus could benefit from being  written in C# code, and thus could be run, producing an output without a judge being called in to make a decision, increasing judicial efficiency.

### 3.4 Research table

| Case Title | Summary | Case result | Summary of the ambiguous language | Quote of said language, with ambiguity highlighted | C# alternative (all LLM produced) | Code Analysis |
|---|---|---|---|---|---|---|
| [10] Rainy Sky SA v. Kookmin Bank | The case is about a dispute over the interpretation of financial guarantees provided by the bank. The main question was whether the guarantees covered refunds if the shipbuilder became insolvent. | The judgment in the case of Rainy Sky SA & Orsd v Kookmin Bank was in favor of the appellants, Rainy Sky SA and others. The United Kingdom Supreme Court allowed the appeal and restored the summary judgment initially granted by the lower court for the Assignee, thereby ordering Kookmin Bank to pay the amounts due under the Bonds. The Supreme Court found that the construction of the Bonds, as proposed by the Buyers, which covered refunds due under Article XII.3 of the Contracts, was consistent with | The highlighted ambiguous phrase was interpreted in 2 ways: 1. The buyers argue that the bank should refund all pre-delivery payments especially if the shipbuilder becomes insolvent. 2. the bank believes that the language only covers refunds for contract cancellations etc. This confusion is due to the unclarity of whether the insolvency situations was supposed to be | "In consideration of your agreement to make the pre-delivery instalments under the Contract and for other good and valuable consideration (the receipt and adequacy of which is hereby acknowledged), we hereby, as primary obligor, irrevocably and unconditionally undertake to pay to you, your successors and assigns, on your first written demand, all such sums due to you under the Contract (or such sums which would have been due to you but for any irregularity, illegality, invalidity or unenforceability in whole or in part of the Contract) PROVIDED THAT the total amount recoverable by you under this Bond | public class RefundGuarantee<br>{<br>private const decimal MaxRefundAmount = 26640000m;<br>private const decimal InterestRate = 0.07m;<br>private const decimal TotalLossInterestRate = 0.10m;<br>public decimal CalculateRefund(DateTime paymentDate, DateTime refundDate, bool isTotalLoss)<br>{<br>decimal principleAmount = MaxRefundAmount;<br>decimal interestRate = isTotalLoss ? TotalLossInterestRate : InterestRate;<br>int daysBetween = (refundDate - paymentDate).Days;<br>decimal interest = principleAmount * interestRate * daysBetween / 365;<br>return principleAmount + interest;<br>}<br>}<br>class Program<br>{<br>static void Main(string[] args)<br>{ | This code produces the output "Calculated Refund Amount: £28,499,690.96." Thus, showing that the AI model believes that a refund should occur. This aligns with the court's decision, where they "ordered Kookmin Bank to pay the amounts due under the Bonds." This shows that in this particular case, it potentially could have been avoided if the contract was drafted and written in a programming language and the executability of it would enable the answer to be produced. |
| | | business common sense and the commercial purpose of the Bonds. | included, and whether merely the limits of what the Bank must cover is detailed only. | shall not exceed US $[26,640,000] ... plus interest thereon at the rate of ... (7%) per annum (or ... (10%) per annum in the case of a Total Loss of the Vessel) from the respective dates of payment by you of such instalments to the date of remittance by telegraphic transfer of such refund." | RefundGuarantee refundGuarantee = new RefundGuarantee();<br>DateTime paymentDate = new DateTime(2023, 1, 1);<br>DateTime refundDate = new DateTime(2023, 12, 31);<br>bool isTotalLoss = false;<br>decimal refundAmount = refundGuarantee.CalculateRefund(paymentDate, refundDate, isTotalLoss);<br>Console.WriteLine($"Calculated Refund Amount: {refundAmount:C}");<br>}<br>} | |
| [11] [Marley vs Rawlings]() | A case involving Mr. and Mrs. Rawling who each signed each other's wills, leading to the issue of circular reference. | In the case of Marley vs Rawlings, initially both the High Court and Court of Appeal ruled against Mr. Marley, deciding that the mistake could not be corrected. However, after being taken to the Supreme Court, they unanimously ruled in favour of Mr. Marley due to Section 20 Article 1 of the 1982 Wills | The ambiguity arises from whether the testator's intention can be satisfied when the wrong will is signed. | The language at issue involves the statutory requirements for a valid will under Sect. 0 of the Wills Act 1837. - "No will shall be valid unless it is in writing, and signed by the testator, or by some other person in his presence and by his | public class Will<br>{<br>public string Content { get; set; }<br>public bool IsSignedByTestator { get; set; }<br>public bool IsSignedByOtherPersonInTestatorPresence { get; set; }<br>public bool IsSignedByOtherPersonByTestatorDirection { get; set; }<br>public bool IsTestatorIntendedToGiveEffect { get; set; }<br>public bool HasClericalError { get; set; }<br>public bool FailedToUnderstandInstructions { get; set; }<br>public Will(string content, bool isSignedByTestator, | If a clerical error was made during rectification, then a design pattern would be breached equaling an infinite loop. The output produced by the code is as follows:<br>Before rectification, is the will valid? True<br>The will has been rectified to carry out the testator's intentions.<br>After rectification, is the will valid? True<br>Current will content: Rectified content to |
| | | Act. (see Section 3.4.1). | | direction, and it appears that the testator intended by his signature to give effect to the will." | bool isSignedByOtherPersonInTestatorPresence,<br>bool isSignedByOtherPersonByTestatorDirection,<br>bool isTestatorIntendedToGiveEffect,<br>bool hasClericalError,<br>bool failedToUnderstandInstructions)<br>{<br>Content = content;<br>IsSignedByTestator = isSignedByTestator;<br>IsSignedByOtherPersonInTestatorPresence = isSignedByOtherPersonInTestatorPresence;<br>IsSignedByOtherPersonByTestatorDirection = isSignedByOtherPersonByTestatorDirection;<br>IsTestatorIntendedToGiveEffect = isTestatorIntendedToGiveEffect;<br>HasClericalError = hasClericalError;<br>FailedToUnderstandInstructions = failedToUnderstandInstructions;<br>}<br>public bool IsValid()<br>{<br>return !string.IsNullOrWhiteSpace(Content) &&<br>(IsSignedByTestator \|\|<br>(IsSignedByOtherPersonInTestatorPresence &&<br>IsSignedByOtherPersonByTestatorDirection)<br>) &&<br>IsTestatorIntendedToGiveEffect;<br>}<br>public void RectifyContent()<br>{<br>if (HasClericalError \|\|<br>FailedToUnderstandInstructions) | carry out the testator's intentions. This shows that the will is viewed as valid due to the clerical error that has occurred, and no clerical errors have occurred in rectification of the will. This aligns with the view expressed by the legal judgement, showing that this contract could have been written in a programming language rather than plain language to increase judicial efficiency. |
| | | | | | {<br>Content = "Rectified content to carry out the testator's intentions.";<br>HasClericalError = false;<br>FailedToUnderstandInstructions = false;<br>Console.WriteLine("The will has been rectified to carry out the testator's intentions.");<br>}<br>}<br>} | |
| [12] Investors Compensation Scheme Ltd v West Bromwich Building Society | The case of Investors Compensation Scheme Ltd v. West Bromwich Building Society revolved around the interpretation of Claim Forms used by the Investors Compensation Scheme (I.C.S.) to take over rights of certain investors. These investors had suffered losses from "Home Income Plans," where home equity was used to secure | In the case of Investors Compensation Scheme Ltd v. West Bromwich Building Society, the appeal was allowed in favor of the appellants, Investors Compensation Scheme Ltd (I.C.S.). The House of Lords held that the claims for damages and compensation had been assigned to I.C.S. The Lords concluded that, on the true construction of the | The ambiguity of this case stems from the poorly drafted parenthesis and the use of the word “abatement.” The phrasing “whether sounding in rescission for undue influence or otherwise” was unclear and misled parties about whether rights were | "I.C.S. agrees that the following claims shall not be treated as a 'Third Party Claim' [as defined in section 4 of this form] for the purposes of this agreement and that the benefits of such claims shall enure to you absolutely: Any claim (whether sounding in rescission for undue influence or otherwise) that you have or may have against the West Bromwich Building Society in which you claim an abatement of sums which you would otherwise have to repay to that Society in respect of sums | using System;<br>public class Claim<br>{<br>public bool IsThirdPartyClaim { get; set; }<br>public double AmountToRepay { get; set; }<br>public double AbatementAmount { get; set; }<br>public void EvaluateClaim()<br>{<br>if (IsEligibleUnderIcsAgreement())<br>{<br>IsThirdPartyClaim = false;<br>AssignBenefitsToClaimant();<br>}<br>}<br>private bool IsEligibleUnderIcsAgreement()<br>{<br>return (HasRescissionOrUndueInfluenceClaim() && | The output from the code is as follows: Benefits of the claim are assigned to the claimant.<br>Is Third Party Claim: False<br>Amount To Repay: 10000<br>Abatement Amount: 2000.<br>This output agrees with the judge’s decision, again showing how a programmable contract would be a more logical choice. |
| | investments that eventually failed.<br><br>The core issue was the interpretation of section 3(b) of the Claim Form, which reserved certain rights against building societies to the investors, while assigning other rights to I.C.S. The investors and I.C.S. had conflicting interpretations of which claims were retained by the investors and which were assigned to I.C.S. | Claim Form, the investors retained the right to claim rescission of their mortgages, but I.C.S. had validly been assigned all claims for compensation and damages. | retained or transferred. | borrowed by you from that Society in connection with the transaction and dealings giving rise to the claim (including interest on any such sums)." | IsAgainstWestBromwichBuildingSociety());<br>}<br>private bool HasRescissionOrUndueInfluenceClaim()<br>{<br>return true;<br>}<br>private bool IsAgainstWestBromwichBuildingSociety()<br>{<br>return true;<br>}<br>private void AssignBenefitsToClaimant()<br>{<br>Console.WriteLine("Benefits of the claim are assigned to the claimant.");<br>}<br>} | |
| [13] R (on the application of Privacy International) v Investigatory Powers Tribunal | In the case of R (Privacy International) v Investigatory Powers Tribunal, the Supreme Court was tasked with determining whether the Investigatory Powers Tribunal's (IPT) decisions could be subject to judicial review and whether section 67(8) of the Regulation of Investigatory Powers Act 2000 (RIPA) allowed for such review to be excluded. Privacy International had challenged computer hacking activities by GCHQ, authorized by thematic warrants under section 5 of the Intelligence Services Act 1994, which they argued were unlawful.<br><br>Section 67(8) of RIPA contains an ouster clause stating that IPT's decisions "shall not be subject to appeal or be liable to be questioned in any | The majority of the Supreme Court found that section 67(8) was not effective in completely ousting judicial review. They determined that there must still be an ability for the High Court to review cases where the Tribunal might have exceeded its jurisdiction, including situations involving errors of law.<br><br>Ultimately, the appeal was allowed, affirming that judicial review remains a crucial mechanism for upholding legal accountability in such instances, and that the | The ambiguity arises from the phrase "including decisions as to whether they have jurisdiction." This language is unclear as it suggests that this exclusion applies to all decisions, including those made in error or outside the jurisdiction, or only to valid decisions. | "Except to such extent as the Secretary of State may by order otherwise provide, determinations, awards, orders and other decisions of the Tribunal (including decisions as to whether they have jurisdiction) shall not be subject to appeal or be liable to be questioned in any court." | public class TribunalDecision<br>{<br>public bool IsAppealable { get; private set; }<br>public bool IsQuestionableInCourt { get; private set; }<br>public TribunalDecision(bool secretaryOfStateOrder)<br>{<br>if (secretaryOfStateOrder)<br>{<br>IsAppealable = true;<br>IsQuestionableInCourt = true;<br>}<br>}<br>public bool HasJurisdiction { get; set; }<br>}<br>public class Program<br>{<br>public static void Main(string[] args)<br>{<br>bool secretaryOfStateOrder = // determine the value based on your logic or input<br>var decision = new TribunalDecision(secretaryOfStateOrder);<br>Console.WriteLine($"Is Appealable: {decision.IsAppealable}");<br>Console.WriteLine($"Is Questionable In Court: {decision.IsQuestionableInCourt}");<br>}<br>} | Ambiguity still arises when translated to code as user must manually set secretaryOfStateOrder to true or false. When scretaryOfStateOrder = true, the output is: Is Appealable: True Is Questionable In Court: True. When secretaryfOfStateOrder = false, the output is: Is Appealable: False Is Questionable In Court: False. This input that is required from the user displays how the contract is ambiguous to the extent that the language cannot be translated into non-user interactive code. |
| | court." This raised the question of whether judicial review was possible, particularly concerning errors of law made by the IPT. | IPT's decisions could still be questioned by the courts where errors of law are alleged. | | | | |
| [14] [Pimlico Plumbers Ltd v Smith]() | The case of Pimlico Plumbers Ltd & Anor v Smith involved Mr. Smith, a plumber who worked | The judgment in Pimlico Plumbers Ltd & Anor v Smith [2018] UKSC 29 was in favor of Mr. Smith, the respondent. | The ambiguity arises primarily from conflicting terms and the nature of Mr. Smith's obligation | "You are an independent contractor... Nothing in this Agreement shall render you an employee, agent or partner of the Company..."<br><br>"The Company shall be under no obligation to offer you work and you | using System;<br>using System.Collections.Generic;<br>public class IndependentContractor<br>{<br>public string Name { get; set; }<br>public bool IsIndependentContractor { get; private set; } = true;<br>public bool HasAcceptedWork { get; set; }<br>public List<string> ToolsAndEquipment { get; private set; } = new List<string>();<br>public IndependentContractor(string name)<br>{<br>Name = name; | The output from the code is:<br>Work is offered to Mr Smith.<br>Mr Smith accepts the work.<br>Mr Smith agrees to provide |
| | for Pimlico Plumbers, claiming he was a "worker" entitled to certain rights under employment law. The UK Supreme Court found in favor of Mr. Smith, affirming that he was a worker under the Employment Rights Act 1996, which allowed him to pursue claims regarding unpaid wages, holiday | The UK Supreme Court ruled that Mr. Smith was a "worker" within the meaning of section 230(3) of the Employment Rights Act 1996. This decision allowed him to proceed with claims related to unlawful wage deductions, unpaid statutory annual leave, and discrimination.<br><br>The court held that Mr. Smith had an obligation of personal performance for Pimlico, and Pimlico | . While the contract states no obligation to either party to offer or accept work, there are provisions that imply structured work expectations which contradict the lack of obligations. Additionally, the concept of “self-employment” is undermined by Mr. Smith’s work conditions, leading to confusion about | shall be under no obligation to accept such work..."<br><br>- "You shall provide the Services for such periods as may be agreed with the Company from time to time."<br><br>- "You warrant … that … you will be competent to perform the work which you agree to carry out."<br><br>- "You shall provide all your own tools, equipment, materials and other items as shall be required for the performance of the Services."<br><br>- "You will have personal liability for the consequences of your services to the Company."<br><br>- "You shall notify the Company … if you are unable to work due to illness or injury." | }<br>public void OfferWork(bool companyOffersWork)<br>{<br>if (companyOffersWork)<br>{<br>Console.WriteLine($"Work is offered to {Name}.");<br>}<br>else<br>{<br>Console.WriteLine($"No work obligation for {Name}.");<br>HasAcceptedWork = false;<br>}<br>}<br>public void AcceptOrDeclineWork(bool acceptWork)<br>{<br>if (acceptWork)<br>{<br>HasAcceptedWork = true;<br>Console.WriteLine($"{Name} accepts the work.");<br>}<br>else<br>{<br>HasAcceptedWork = false;<br>Console.WriteLine($"{Name} declines the work.");<br>}<br>}<br>public void AgreeOnServicePeriod(int days)<br>{<br>Console.WriteLine($"{Name} agrees to provide services for {days} days.");<br>}<br>public void WarrantCompetence()<br>{<br>Console.WriteLine($"{Name} warrants being competent to perform agreed work.");<br>} | services for 5 days.<br>Mr Smith warrants being competent to perform agreed work.<br>Mr Smith provides own tools: tools.<br>Independent Contractor: Mr Smith, Tools: tools, Is Independent: True, Accepted Work: True. To break it down, the output states that Mr. Smith has been offered work, aligning with the judgement. The detail of the tools supports the independent contractor perspective. Overall, the code does not match the legal judgement as the output suggests from |
| | pay, and discrimination. | was not just a client or customer. The tribunal had been entitled to conclude that Mr. Smith had the status of a worker and not an independent contractor, despite the contractual language suggesting otherwise. | whether Mr. Smith is characterised as a client or customer under the worker definition. | - "You are an independent contractor of the Company, in business on your own account." | public void ProvideTools(params string[] tools)<br>{<br>ToolsAndEquipment.AddRange(tools);<br>Console.WriteLine($"{Name} provides own tools: {string.Join(", ", tools)}.");<br>}<br>public void NotifyIllness(bool isIll)<br>{<br>if (isIll)<br>{<br>Console.WriteLine($"{Name} notifies the company of inability to work due to illness.");<br>}<br>}<br>public override string ToString()<br>{<br>return $"Independent Contractor: {Name}, Tools: {string.Join(", ", ToolsAndEquipment)}, " +<br>$"Is Independent: {IsIndependentContractor}, Accepted Work: {HasAcceptedWork}";<br>}<br>}<br>class Program<br>{<br>static void Main()<br>{<br>var contractor = new IndependentContractor("John Doe");<br>contractor.OfferWork(true);<br>contractor.AcceptOrDeclineWork(true);<br>contractor.AgreeOnServicePeriod(5);<br>contractor.WarrantCompetence();<br>contractor.ProvideTools("Hammer", "Saw", "Drill");<br>contractor.NotifyIllness(false);<br>Console.WriteLine(contractor);<br>}<br>} | an independent contractor perspective, whilst the legal judgement establishes him as a worker. This highlights both possible flaws in the LLM and the ambiguity of the contract being so high that the LLM has interpreted it alternatively. |
| | | | | | | |

| | | | | | | | |
|---|---|---|---|---|---|---|---|
| [15] Uber BV v Aslam | The primary focus of this case is whether Uber drivers are considered “workers” under employment law, which would entitle them to more rights, or whether they are independent contractors as Uber contested. | The UK Supreme Court ruled unanimously in favor of the respondents, Yaseen Aslam and others, affirming that Uber drivers are "workers" for the purposes of employment legislation. This means that the drivers are entitled to basic employment rights such as the national minimum wage, paid annual leave, and protection against detrimental treatment for whistleblowing. The Court agreed with the findings of the Employment Appeal Tribunal and the majority of the Court of Appeal, rejecting Uber's contention that drivers were independent contractors who worked for themselves. Key factors supporting the decision included Uber's significant control over how drivers provided their services, the set | The language in this contract is ambiguous as it attempts to establish a relationship where Uber is not a party to the business relationship between drivers and users, whilst simultaneously exercising significant control over the driver’ operations. The clause suggests that Uber does not control the drivers, whilst imposing performance standards on the drivers. | - Clause 2.3 of the Services Agreement: "Customer acknowledges and agrees that Customer’s provision of Transportation Services to Users creates a legal and direct business relationship between Customer and the User, to which neither Uber [BV] nor any of its Affiliates in the Territory is a party."<br>- Clause 2.4 of the Services Agreement:<br>- "Uber and its Affiliates in the Territory do not, and shall not be deemed to, direct or control Customer or its Drivers generally or in their performance under this Agreement specifically, including in connection with the operation of Customer’s business, the provision of Transportation Services, the acts or omissions of Drivers, or the operation and maintenance of any Vehicles."<br>- Clause 4.1 of the Services Agreement:<br>- "Customer: (i) appoints Uber [BV] as Customer’s limited payment collection | using System;<br>public class UberContract<br>{<br>public bool IsUberPartyToBusinessRelationship { get; private set; } = false;<br>public bool DoesUberControlDriverPerformance { get; private set; } = false;<br>public void AcknowledgeBusinessRelationship()<br>{<br>IsUberPartyToBusinessRelationship = false;<br>}<br>public void AcknowledgeControl()<br>{<br>DoesUberControlDriverPerformance = false;<br>}<br>public void AppointPaymentAgent()<br>{<br>bool isUberPaymentAgent = true;<br>bool isPaymentToUberConsideredDirect = true;<br>Console.WriteLine("Uber appointed as payment agent.");<br>}<br>}<br>public class Program<br>{<br>public static void Main(string[] args)<br>{<br>UberContract contract = new UberContract();<br>contract.AcknowledgeBusinessRelationship();<br>Console.WriteLine($"Is Uber a party to the business relationship? {contract.IsUberPartyToBusinessRelationship}");<br>); | The output from the code is as follows: Is Uber a party to the business relationship? False Does Uber control driver performance? False Uber appointed as payment agent. The judgement and output of the code do not match, the legal judgement classifies drivers as workers whilst the code contradicts the courts findings. In |
| | | fare structure, and the monitoring and management of drivers' performance through the app. | | agent solely for the purpose of accepting the Fare … on behalf of the Customer via the payment processing functionality facilitated by the Uber Services; and (ii) agrees that payment made by User to Uber [BV] shall be considered the same as payment made directly by User to Customer." | contract.AcknowledgeControl();<br>Console.WriteLine($"Does Uber control driver performance? {contract.DoesUberControlDriverPerformance}");<br>contract.AppointPaymentAgent();<br>}<br>} | this case I conclude the LLM to be inaccurate as the factual evidence has been contradicted. |
| [16] Arnold v Britton | This case related to the interpretation of the service charge provisions. The lease includes a clause indicating that lessees must pay a "proportionate part of the expenses" alongside | Ultimately, the majority ruled that the lessor was entitled to charge the fixed sums as prescribed by the service charge clauses in the leases, resulting in the leases being enforceable as initially interpreted to have a service charge that increased annually at a compound rate of 10%. This | The ambiguity revolves around whether the yearly sum of £90, adjusted by 10% every three years, is meant to be a fixed amount or a cap on the service charge. The language suggests a potential conflict between a variable | "To pay to the Lessors without any deductions in addition to the said rent a proportionate part of the expenses and outgoings incurred by the Lessors in the repair maintenance renewal and the provision of services hereinafter set out the yearly sum of Ninety | using System;<br>public class Lease<br>{<br>private const decimal InitialServiceCharge = 90m;<br>private const decimal IncreaseRate = 0.10m;<br>private const int PeriodYears = 3;<br>private const decimal VATRate = 0.20m;<br>public int Year { get; private set; }<br>public decimal AdditionalRent { get; private set; }<br>public decimal ExpensesIncurred { get; private set; }<br>public Lease(int startYear, decimal expensesIncurred)<br>{<br>Year = startYear;<br>AdditionalRent = InitialServiceCharge;<br>ExpensesIncurred = expensesIncurred;<br>}<br>public void CalculateYearlyCharge()<br>{<br>int yearsElapsed = Year - 1974;<br>int periodsElapsed = yearsElapsed / PeriodYears;<br>AdditionalRent = InitialServiceCharge * (decimal)Math.Pow(1 + IncreaseRate, periodsElapsed); | The output produced can be seen below this table. (first 10 years). The output from the code and the legal judgement align as the code output reflects the legal judgement of increasing legal charges. |
| | a fixed yearly sum. The issue revolved around whether this clause set a fixed amount or acted as a cap on service charges, due to the ambiguity surrounding the combination of the language about expenses. | decision left the lessees in a potentially financially burdensome position since the charges could become very substantial in the later years of the lease. | expense and a capped amount. | Pounds and value added tax (if any) for the first three years of the term hereby granted increasing thereafter by Ten Pounds per Hundred for every subsequent Three year period or part thereof." | decimal proportionatePart = CalculateProportionatePart(ExpensesIncurred, chargeWithVAT);<br>Console.WriteLine($"Year: {Year} - Yearly Charge: {chargeWithVAT:C} - Proportionate Part: {proportionatePart:C}");<br>}<br>private decimal CalculateProportionatePart(decimal totalExpenses, decimal chargeWithVAT)<br>{<br>int numberOfUnits = 91; //hypothetical<br>decimal proportionatePart = totalExpenses / numberOfUnits;<br>proportionatePart = Math.Min(proportionatePart, chargeWithVAT);<br>return proportionatePart;<br>}<br>public void NextYear()<br>{<br>Year++;<br>} }<br>public class Program<br>{<br>public static void Main(string[] args)<br>{<br>decimal totalExpensesIncurred = 2000m;<br>Lease lease = new Lease(1974, totalExpensesIncurred);<br>for (int i = 0; i < 99; i++) // Assuming a 99-year lease<br>{<br>lease.CalculateYearlyCharge();<br>lease.NextYear();<br>}}} | |

Code Output:

Year: 1974 - Yearly Charge: £108.00 - Proportionate Part: £21.98
Year: 1975 - Yearly Charge: £108.00 - Proportionate Part: £21.98
Year: 1976 - Yearly Charge: £108.00 - Proportionate Part: £21.98
Year: 1977 - Yearly Charge: £118.80 - Proportionate Part: £21.98
Year: 1978 - Yearly Charge: £118.80 - Proportionate Part: £21.98

Year: 1979 - Yearly Charge: £118.80 - Proportionate Part: £21.98
Year: 1980 - Yearly Charge: £130.68 - Proportionate Part: £21.98
Year: 1981 - Yearly Charge: £130.68 - Proportionate Part: £21.98
Year: 1982 - Yearly Charge: £130.68 - Proportionate Part: £21.98
Year: 1983 - Yearly Charge: £143.75 - Proportionate Part: £21.98
Year: 1984 - Yearly Charge: £143.75 - Proportionate Part: £21.98

| | | | | | | |
|---|---|---|---|---|---|---|
| [17] BCCI v Ali | The case constitutes mainly of the broad language of the release, and whether "all or any claims" encompasses claims for damage due to the stigma of having to work for a corrupt bank. | The appeal was dismissed, allowing Mr. Naeem to pursue his claim for stigma damages.<br><br>This decision reflects the principle that a party is unlikely to have intended to release claims of which they were unaware unless expressly stated in the | The language is ambiguous due to the use of broad and general terms, which can be interpreted to include claims that were unknown or unforeseen at the time. | "The applicant agrees to accept the terms set out in the documents attached in full and final settlement of all or any claims whether under statute, common law or in equity of whatsoever nature that exist or may exist and, in particular, all or any claims rights or applications of whatsoever nature that the applicant has or may have or has made or could make in | using System;<br>public class SettlementAgreement<br>{<br>public bool AcceptTerms { get; set; }<br>public bool HasExistingClaims { get; set; }<br>public bool HasPotentialClaims { get; set; }<br>public bool HasMadeIndustrialTribunalClaims { get; set; }<br>public bool CouldMakeIndustrialTribunalClaims { get; set; }<br>public bool HasPensionSchemeRights { get; set; }<br>public void ExecuteSettlement()<br>{<br>if (AcceptTerms)<br>{<br>if ((HasExistingClaims \|\| HasPotentialClaims \|\| HasMadeIndustrialTribunalClaims \|\| CouldMakeIndustrialTribunalClaims) &&<br>ClaimsAreForeseeable())<br>{<br>SettleClaims();<br>}<br>if (HasPensionSchemeRights)<br>{<br>PreservePensionSchemeRights();<br>}<br>HandleUnknownClaims();<br>}<br>}<br>private void SettleClaims() | The output from the code is as follows: Claims have been settled. Pension scheme rights preserved. Handling unknown claims. The legal judgement and the output do not align, indicating either an LLM code error or ambiguity not being translated into code due to the |

| | | | | | | | |
|---|---|---|---|---|---|---|---|
| | | release's language. | | or to the industrial tribunal, except the applicant's rights under the respondent's pension scheme." | {<br>Console.WriteLine("Claims have been settled.");<br>}<br>private void PreservePensionSchemeRights()<br>{<br>Console.WriteLine("Pension scheme rights preserved.");<br>}<br>private bool ClaimsAreForeseeable()<br>{<br>return true; // Placeholder for real logic<br>}<br>private void HandleUnknownClaims()<br>{<br>Console.WriteLine("Handling unknown claims.");<br>}<br>}<br>public class Program<br>{<br>public static void Main(string[] args)<br>{<br>SettlementAgreement agreement = new SettlementAgreement<br>{<br>AcceptTerms = true,<br>HasExistingClaims = true,<br>HasPotentialClaims = false,<br>HasMadeIndustrialTribunalClaims = false,<br>CouldMakeIndustrialTribunalClaims = true,<br>HasPensionSchemeRights = true<br>};<br>agreement.ExecuteSettlement();<br>} | unclear language. |
| [18] Lloyds TSB Foundation for Scotland v Lloyds Banking Group plc | The case centered around the interpretation of a covenant in a deed. The dispute arose over the calculations of payments the bank was obligated to make to the Foundation, specified as a percentage of the pre-tax profits or a fixed minimum amount. With there being varying interpretations of "group profit before taxation." | The Supreme Court concluded that the deed should be understood in the context of the legal and accounting principles prevailing at the time it was executed. These principles focused on realized profits rather than unrealized gains. The court found it unthinkable at the time the deed was made that unrealized profits would be included, and such changes were not foreseen.<br><br>Thus, the appeal was allowed, restoring the decision of the Lord Ordinary to grant decree of absolvitor and agreeing with Lloyds Banking Group's interpretation. | The ambiguous language stems from the way in which "Pre-Tax Profits" and "Pre-Tax Loss" are to be calculated considering unforeseen accounting practices. As the original contract did not anticipate future changes in accounting standards. Therefore, the inclusion of an unrealised "gain on acquisition" created substantial ambiguity.<br><br>The code reflects the ambiguity in whether "(a) one-third of 0.1946% of Pre-Tax Profits (after deducting Pre-Tax Losses)" should be interpreted as directly applying the percentage to the profit difference or involves another operation beyond this calculation | "Lloyds Bank covenanted to pay the Foundation the greater of "(a) an amount equal to one-third of 0.1946 per cent of the Pre-Tax Profits (after deducting Pre-Tax Losses)" for the relevant Accounting Reference Periods and "(b) the sum of £38,920". | public class PaymentCalculator<br>{<br>public decimal CalculatePayment(decimal preTaxProfits, decimal preTaxLosses)<br>{<br>decimal calculatedPayment = (preTaxProfits - preTaxLosses) * (0.1946m / 3);<br>decimal minimumPayment = 38920m;<br>return Math.Max(calculatedPayment, minimumPayment);<br>}<br>}<br>public class Program<br>{<br>public static void Main(string[] args)<br>{<br>PaymentCalculator calculator = new PaymentCalculator();<br>decimal preTaxProfits = 100000m; // Example pre-tax profits<br>decimal preTaxLosses = 20000m; // Example pre-tax losses<br>decimal payment = calculator.CalculatePayment(preTaxProfits, preTaxLosses);<br>System.Console.WriteLine($"The payment to the Foundation is: £{payment}");<br>}<br>} | The code output: The payment to the Foundation is: £38920. The code output provides a monetary amount which is reliant on input data from the user. This is not directly matched to the legal judgement, and to provide a better match the output would need to be contextualised. This result could either be from the LLM error, or from ambiguity that resulted in |
| | | | | | | misinterpretation. |
| [19] MonSolar IQ Ltd v Woden Park Ltd | In the case "Monsolar IQ Ltd v Woden Park Ltd [2021] EWCA Civ 961," the Court of Appeal dealt with a rent review clause dispute. The lease, intended for a solar farm, contained a formula that led to exponential rent increases. MonSolar argued for a correction under the Chartbrook principle, which allows for the correction of clear drafting mistakes.<br><br>The High Court found the formula to be irrational and arbitrary, not aligning with the intended purpose of reflecting changes in the Retail Prices Index (RPI). It was determined that the mistake was due to mixing two indexing | The case "Monsolar IQ Ltd v Woden Park Ltd [2021] EWCA Civ 961" involved an appeal by Woden Park Ltd regarding a rent review clause in a lease agreement with MonSolar IQ Ltd. The Court of Appeal dismissed the appeal, agreeing with the High Court's decision that there was a drafting mistake in the lease's rent review formula, which led to an exponential rent increase. The court held that it was clear both that a mistake had been made and how it should be corrected, resulting in a correction to ensure the rent | The ambiguity arises from the exponential effects the Formula has on rent increases. When applied literally, the Formula results in the rent escalating at a rate significantly higher that the RPI. | "The Rent payable under this Lease will be reviewed in accordance with this paragraph 3 on each of the Review Dates and such Rent payable from and including each such Review Date shall be the Revised Rent which shall be calculated as follows:<br><br>Revised Rent = Rent payable prior to the Review Date (disregarding any suspension of Rent) x Revised Index | using System;<br>public class RentCalculator<br>{<br>figures<br>public double CalculateRevisedRent(double previousRent, double revisedIndexFigure, double baseIndexFigure)<br>{<br>figure<br>if (baseIndexFigure == 0)<br>{<br>throw new DivideByZeroException("Base index figure must not be zero.");<br>}<br>return previousRent * (revisedIndexFigure / baseIndexFigure);<br>}<br>}<br>public class Program<br>{<br>public static void Main(string[] args)<br>{<br>RentCalculator rentCalculator = new RentCalculator();<br>double previousRent = 1000.0; // Example Rent amount before the review date<br>double revisedIndexFigure = 105.0; // Example revised index figure<br>double baseIndexFigure = 100.0; // Example base index figure | The output is as follows: The revised rent is: 1050. The output from the code matches the legal judgement to an extent as the legal judgement does not state a specific revised rent figures, however, does indicate that the revised rent will be calculated in an alternate manner. |
| | methods, leading to extreme rent hikes. | increased in line with RPI changes annually. | | Figure / Base Index Figure" | try<br>{<br>double revisedRent =<br>rentCalculator.CalculateRevisedRent(previousRent, revisedIndexFigure, baseIndexFigure);<br>Console.WriteLine($"The revised rent is: {revisedRent}");<br>}<br>catch (DivideByZeroException ex)<br>{<br>Console.WriteLine($"Error: {ex.Message}");<br>}<br>}<br>} | |
| [20] European Professional Club Rugby v RDA Television LLP | The main conflict arose when EPCR postponed the playoff stages of the 2019–2020 competitions due to COVID-19, which constituted a force majeure event. RDA asserted that this event hindered EPCR's performance under the | In the case of European Professional Club Rugby v RDA Television LLP [2022] EWHC 50 (Comm), the result was in favour of RDA Television LLP. The court | The force majeure clause is found in Clause 26 of Schedule 6 of the Media Rights Agreement (MRA) between EPCR and RDA. This clause outlines the circumstances in which a party may be relieved | "… any circumstances beyond the reasonable control of a party affecting the performance by that party of its obligations under this Agreement including inclement weather conditions, | using System;<br>public class RentCalculator<br>{<br>public enum ForceMajeureEvent<br>{<br>InclementWeather,<br>Fire,<br>Storm,<br>Flood,<br>Lightning,<br>Earthquake,<br>Explosion,<br>ActsOfPublicEnemy,<br>Terrorism,<br>War,<br>MilitaryOperations,<br>Insurrection,<br>Sabotage,<br>CivilDisorder,<br>Epidemic, | They use the phrase “other” in the force majeure clause. This is an example of “Catch-All” (anti-pattern in software development), as the use of “other” spans many options. The output from the code is: Is Force Majeure |
| | contract, specifically its fundamental obligation to stage the competitions, thus allowing RDA to terminate the MRA under a force majeure clause.<br><br>EPCR contended that the force majeure clause did not give RDA a right to terminate the contract and accused RDA of wrongful repudiation. EPCR argued that RDA should have used reasonable endeavors to mitigate the consequences of the pandemic, as required by the contract, | determined that RDA was entitled to terminate the Media Rights Agreement (MRA) with EPCR (European Professional Club Rugby) under the force majeure provisions of the agreement. The court found that the COVID-19 pandemic constituted a force majeure event that hindered EPCR's performance obligations under the MRA. | of its obligations due to events beyond its reasonable control and allows for contract termination if such events prevent, hinder, or delay performance for a specified period. | serious fir~e, storm, flood, lightning, earthquake, explosion, acts of a public enemy, terrorism, war, military operations, insurrection, sabotage, civil disorder, epidemic, embargoes and labour disputes of a person other than such party." | Embargo,<br>LabourDispute,<br>Other // Represents unforeseen events not specifically listed<br>}<br>public class ForceMajeureChecker<br>{<br>public bool IsForceMajeureEvent(ForceMajeureEvent eventType, bool isBeyondControl, bool isThirdPartyLabourDispute = false)<br>{<br>if (!isBeyondControl)<br>{<br>return false;<br>}<br>switch (eventType)<br>{<br>case ForceMajeureEvent.InclementWeather:<br>case ForceMajeureEvent.Fire:<br>case ForceMajeureEvent.Storm:<br>case ForceMajeureEvent.Flood:<br>case ForceMajeureEvent.Lightning:<br>case ForceMajeureEvent.Earthquake:<br>case ForceMajeureEvent.Explosion:<br>case ForceMajeureEvent.ActsOfPublicEnemy:<br>case ForceMajeureEvent.Terrorism:<br>case ForceMajeureEvent.War:<br>case ForceMajeureEvent.MilitaryOperations:<br>case ForceMajeureEvent.Insurrection:<br>case ForceMajeureEvent.Sabotage:<br>case ForceMajeureEvent.CivilDisorder:<br>case ForceMajeureEvent.Epidemic:<br>case ForceMajeureEvent.Embargo:<br>case ForceMajeureEvent.Other:<br>return true;<br>case ForceMajeureEvent.LabourDispute:<br>return isThirdPartyLabourDispute;<br>default: | Event: True. This is because “other” encompasses all events but would not be reflective of real-life as not all events are encompassed by force majeure clauses, thus this links to there being a fundamental issue with the drafting of force majeure clauses which is addressed with a possible resolution later in the paper. |
| | instead of terminating the agreement. | Consequently, RDA was not liable for wrongful termination, and EPCR's claim for damages due to alleged wrongful repudiation was dismissed. | | | return false;<br>}<br>}<br>}<br>}<br>public class Program<br>{<br>public static void Main()<br>{<br>RentCalculator.ForceMajeureChecker checker = new RentCalculator.ForceMajeureChecker();<br>var eventType = RentCalculator.ForceMajeureEvent.Fire;<br>bool isBeyondControl = true;<br>bool result = checker.IsForceMajeureEvent(eventType, isBeyondControl);<br>Console.WriteLine($"Is Force Majeure Event: {result}");<br>}<br>} | |

### 3.4.1 Circular reference and infinite loop

In the case of Marley vs Rawlings, initially both the High Court and Court of Appeal ruled against Mr. Marley, deciding that the mistake could not be corrected. However, after being taken to the Supreme Court, they unanimously ruled in favour of Mr. Marley due to Section 20 Article 1 of the 1982 Wills Act:

"(1) If a court is satisfied that a will is so expressed that it fails to carry out the testator's intentions, in consequence—
(a) of a clerical error; or
(b) of a failure to understand his instructions,
it may order that the will shall be rectified so as to carry out his intentions."

However, interestingly, the Act does not specify what happens in the event of a clerical error in the rectification of the will. Thus, leading to the form of infinite loop or circular reference (as if a clerical error is made in the rectification then the will can be rectified and so on). Another step that was taken to explore the Marley vs Rawlings case further, due to its complex cyclical nature was providing the LLM with the entire of Section 20 of the Wills Act, and to see how it changes the initial C# code it produced. This experiment was to explore if the Supreme Court had missed anything in the Wills Act that would change the result of the trial. Interestingly, the AI chose to focus in on Article 2 which reads as follows:

"(2) An application for an order under this section shall not, except with the permission of the court, be made after the end of the period of six months from the date on which representation with respect to the estate of the deceased is first taken out."

This was intriguing to us from an outside perspective as the article suggests that there is a time limit, except for when the court deems it irrelevant. This itself is read as a waste of legislation in our view, on the basis that the court will always be the one interpreting this legislation so far as its relevance in forming judgement. Whilst perhaps a helpful guide, this sort of drafting offers no value to a deterministic compiler for automatic execution. Further to this, we questioned the LLM about how the article has no impact on the legislation, other than to just be a guide, and therefore when translated into code, what code based anti-pattern does this legal drafting link to? The AI produced two answers:

1. **Design by Committee**: This occurs when decisions (including specifications and constraints) are introduced through a complex, often bureaucratic decision-making process. In software, it leads to overly complicated and often unnecessary features or constraints that don't serve a clear purpose for the end user. The inclusion of a time limit that can be overridden by court discretion without clear guidance on when it wouldn't be is akin to adding complexity without benefit. It's a guideline that has no real impact since the ultimate decision lies with the court.

2. **Speculative Generality**: This happens when code (or policy, in this case) is added in anticipation of future scenarios that may not actually materialize. The inclusion of a time limit with broad exceptions relies on speculation about its usefulness and real-world applicability. It's meant to address potential future decisions the court might face, but without a clear and enforceable application, it becomes speculative.

Overall, from a non-legal perspective, this time limit serves as a guideline which lacks enforceability thus leading to it having minimal substantial impact. This links to the overall investigation of this paper into ambiguity in legal contracts and shows how even Acts can have irrelevant/broad phrasing, which can be deemed unnecessary. This unnecessary language further backs up the point of this thesis, as if this Act was to be written in a programming language, then this redundant legislation would not be present as it breaches design patterns in code.

### 3.4.2 Force Majeure Clauses and the "Catch-All" Anti-Pattern

At first glance the "Catch-All" anti-pattern and force majeure clauses belong to two completely different domains – software development and legal work. However when converting a Force Majeure clause [Source A] to C# code [Source B] the following occurs:

**Source A:**

*"Neither party shall be liable for any failure or delay in the performance of its obligations under this Agreement if such failure or delay is caused by or results from events beyond the reasonable control of such party, including but not limited to, acts of God; natural disasters; fire; flood; earthquake; storm; war; civil commotion; acts of terrorism; governmental action; labor disputes, strikes, or lockouts; pandemic or epidemic; shortage of or inability to obtain materials, equipment, or transportation; power failure or breakdown of machinery; or any other similar events (each, a "Force Majeure Event").*

*The party affected by a Force Majeure Event shall promptly notify the other party in writing upon becoming aware of such event, providing reasonable details of the occurrence, its expected duration, and its anticipated impact on the performance of obligations hereunder.*

*During the period in which performance is delayed or prevented by a Force Majeure Event, the obligations of the affected party shall be suspended to the extent affected by the Force Majeure Event and the affected party shall use reasonable efforts to mitigate the impact of the Force Majeure Event and resume performance as soon as reasonably practicable.*

*If a Force Majeure Event continues for a period exceeding [number] days, either party may terminate this Agreement upon written notice to the other party, without any liability other than obligations accrued up to the date of termination."*

**Source B:**

```
using System;
namespace ContractManagement
{
public class ForceMajeureClause
{
public enum ForceMajeureEvent
{
ActsOfGod,
NaturalDisasters,
Fire,
Flood,
Earthquake,
Storm,
```

```
War,
CivilCommotion,
ActsOfTerrorism,
GovernmentalAction,
LaborDisputes,
Strikes,
Lockouts,
Pandemic,
Epidemic,
ShortageOfMaterials,
EquipmentFailure,
TransportationFailure,
PowerFailure,
MachineryBreakdown,
Other
}
public class ContractParty
{
public string Name { get; set; }
public void Notify(ContractParty otherParty, ForceMajeureEvent eventOccurred, string details)
{
Console.WriteLine($"{Name} notifies {otherParty.Name} of {eventOccurred}: {details}");
}
public void SuspendObligations(TimeSpan duration)
{
Console.WriteLine($"{Name} obligations suspended for {duration.TotalDays} days.");
}
public void ResumePerformance()
{
Console.WriteLine($"{Name} resumes performance of obligations.");
}
}
public static void HandleForceMajeureEvent(ContractParty affectedParty, ContractParty otherParty, ForceMajeureEvent
eventOccurred, TimeSpan maxDuration)
{
affectedParty.Notify(otherParty, eventOccurred, "Details of the event.");
// Example of suspending obligations
affectedParty.SuspendObligations(maxDuration);
if (maxDuration.TotalDays > 30) // Replace with appropriate logic for [number] days
{
TerminateAgreement(affectedParty, otherParty);
}
else
{
affectedParty.ResumePerformance();
}
}
public static void TerminateAgreement(ContractParty party1, ContractParty party2)
{
Console.WriteLine($"Either {party1.Name} or {party2.Name} may terminate the agreement due to prolonged Force Majeure
Event.");
}
}
class Program
```

```
{
static void Main(string[] args)
{
var partyA = new ForceMajeureClause.ContractParty { Name = "Party A" };
var partyB = new ForceMajeureClause.ContractParty { Name = "Party B" };
ForceMajeureClause.HandleForceMajeureEvent(partyA, partyB, ForceMajeureClause.ForceMajeureEvent.NaturalDisasters,
TimeSpan.FromDays(40));
}
}
}
```

Contrasting the highlighted sections in Source A and Source B, one can clearly see the inclusion of the term “other” both in the code and plain text. This force majeure clause, when converted to C# code, results in an anti-pattern: "Catch-All". This phrase “other” lacks clarity, therefore this displays the ambiguity evident in the contract and this suggests that it should not have been drafted that way. This maps to the anti-pattern “Catch-All” which denotes when a general-purpose condition attempts to cover any undefined or unexpected scenario, incredibly similar to the definition of a force majeure clause itself. However, this approach can lead to inadvertently encompassing cases that do not truly belong. When the GPT4 LLM: [21] in Jylo was queried with the question that ‘the word ’other’ was too broad and vague and therefore how would it rewrite the C# code to remove this ambiguity?’ it produced two solutions. Firstly, it proposed a method of using free-form text description and allowing users to specify events as needed. This would reduce ambiguity as all events under the clause could be stored in a separate text document, and therefore no ambiguity would arise over whether an event is encompassed in the force majeure clause. The code that it produced is below:

```
public class ForceMajeureClause
{
// Other existing code...
public static void HandleCustomEvent(ContractParty affectedParty, ContractParty otherParty, string customEventDescription,
TimeSpan duration)
{
Console.WriteLine($"Custom Force Majeure Event: {customEventDescription}");
affectedParty.Notify(otherParty, ForceMajeureEvent.Other, customEventDescription);
// Logic for handling the event
affectedParty.SuspendObligations(duration);
if (duration.TotalDays > 30)
{
TerminateAgreement(affectedParty, otherParty);
}
else
{
affectedParty.ResumePerformance();
}
}
}
```

The second proposition of the LLM was a more structured approach, allowing events to be checked to see if they have already been registered. This more defined and structured approach can be seen below:

```
using System.Collections.Generic;
```

```
public class ForceMajeureManager
{
private static HashSet<string> CustomEvents = new HashSet<string>();
public static void RegisterCustomEvent(string eventName)
{
if (!string.IsNullOrWhiteSpace(eventName))
{
CustomEvents.Add(eventName);
Console.WriteLine($"Registered custom event: {eventName}");
}
}
public static bool IsCustomEventRegistered(string eventName)
{
return CustomEvents.Contains(eventName);
}
public static void HandleRegisteredEvent(ContractParty affectedParty, ContractParty otherParty, string eventName, TimeSpan
duration)
{
if (IsCustomEventRegistered(eventName))
{
Console.WriteLine($"Handling registered event: {eventName}");
affectedParty.Notify(otherParty, ForceMajeureEvent.Other, eventName);
affectedParty.SuspendObligations(duration);
if (duration.TotalDays > 30) // Customize the threshold as needed
{
TerminateAgreement(affectedParty, otherParty);
}
else
{
affectedParty.ResumePerformance();
}
}
else
{
Console.WriteLine($"Event {eventName} is not recognized as a registered custom event.");
}
}
}
// Usage of the structured approach
public class Program
{
static void Main(string[] args)
{
var partyA = new ForceMajeureClause.ContractParty { Name = "Party A" };
var partyB = new ForceMajeureClause.ContractParty { Name = "Party B" };
ForceMajeureManager.RegisterCustomEvent("Meteor Strike");
ForceMajeureManager.HandleRegisteredEvent(partyA, partyB, "Meteor Strike", TimeSpan.FromDays(10));
}
}
```

Consequently, as suggested by the AI, to reduce the ambiguity prevalent in force majeure clauses, the evident step would be to remove the use of “other” and replace this phrasing with a solution that encompasses all wanted events. Such a solution is explored further in Section 3.4.2.1.

### 3.4.2.1 Force Majeure Clause resolution to reduce broadness

As previously iterated, the opinion of Force Majeure clauses, as evident through the thesis, is that they are too broad with the inclusion of “others” or “including” in the clause, lacking clarity, which is what this thesis aims to address. Thus, as a solution we have aimed to draft a theory which removes the “other” phrasing in force majeure clauses, leading to reduced ambiguity in the contract. Our theory involves asking an LLM to extract, from the force majeure clause, all the elements of the clause, discounting “other”. Then the LLM was asked to generate a list of 20 events which were ranked 10-1 on how similar they were to the listed events in the force majeure clause, and secondly were ranked 10-1 on how significant their impact was, from the perspective of the AI model. The results were as follows, starting with the Force Majeure clause inputted (identical to Source A):

"Neither party shall be liable for any failure or delay in the performance of its obligations under this Agreement if such failure or delay is caused by or results from events beyond the reasonable control of such party, including but not limited to, acts of God; natural disasters; fire; flood; earthquake; storm; war; civil commotion; acts of terrorism; governmental action; labor disputes, strikes, or lockouts; pandemic or epidemic; shortage of or inability to obtain materials, equipment, or transportation; power failure or breakdown of machinery; or any other similar events (each, a "Force Majeure Event")."

This force majeure clause produced the following table of similar events:

| Event | Similarity Ranking | Potential Impact Ranking |
|---|---|---|
| Explosive Volcanic Eruption | 10 | 9 |
| Tsunami | 10 | 10 |
| Nuclear Accident | 9 | 10 |
| Cyberattack on Critical Infrastructure | 8 | 9 |
| Riot or Large-Scale Protest | 8 | 8 |
| Severe Supply Chain Disruption due to global events | 7 | 8 |
| Severe Recession causing widespread disruption | 7 | 7 |
| Large-Scale Epidemic (not global) | 7 | 8 |
| Global Trade War | 6 | 7 |
| Major Transportation Strike | 6 | 6 |
| Industrial Accident leading to explosion | 6 | 7 |
| Severe Weather Event (e.g., blizzard) | 5 | 5 |
| Wildfire | 5 | 6 |
| Political Coup causing civil disorder | 5 | 8 |
| Currency Crisis impacting trade | 4 | 6 |
| Major Infrastructure Failure | 4 | 7 |
| Regulatory Change due to unforeseen events | 3 | 5 |
| Technological Failure (e.g., satellite disruption) | 3 | 4 |
| Major Theft or Vandalism impacting operations | 2 | 4 |
| Technological Change affecting business landscape | 1 | 3 |

The next step proposed would be to filter out the events included in the Force Majeure clause, which would be only ones that had a ranking greater than or equal to 7 (or any chosen number) for similarity compared to the original events in the clause or those that had a ranking greater than or equal to 7 (or any chosen number) for the impact/severity ranking. This Boolean logic can also be expressed as C# code, which would read as the following:

```
int similarityRanking = 8; // Example value, can be set according to your data
int impactsRanking = 6; // Example value, can be set according to your data
if (similarityRanking >= 7 || impactsRanking >= 7)
{
Console.WriteLine("The event is in the force majeure clause.");
}
else
{
Console.WriteLine("The event is not in the force majeure clause.");
}
```

This in theory would be executed in a legal contract by having the AI prompt as a replacement of the Force Majeure clause. The input would be the event along with the AI prompt, which would then subsequently output a ‘ranking’ out of 10 for the similarity to other events and the severity. By implementing this approach, legal practitioners can achieve greater precision in force majeure clauses, leading to fewer disputes and clear contractual obligations.

# 4 Evaluation

## 4.1 Data and Research analysis

From research conducted, we concluded that there were three main types of ambiguity found in contracts:

1. Ambiguity through a certain phrase or sentence.

2. Ambiguity through the whole extract.

3. Ambiguity through the absence of mentioning certain phrases or terminology.

Additionally, when compiling our research, we selectively chose to not include case-law that contained ambiguity in the results table (Section 3.4) as information was missing from the contract, and therefore this does not translate into executable code, as some variables would not have a value. However, for the purpose of the research into types of ambiguity, the results are still included above and the table detailing certain cases with ambiguity through absence can be seen in Section 4.1.1.

Another element we were assessing in our research was whether the output of the result by the code aligned with the judge's decision. Consequently, like the table above, we group the research into three groups:

1. Judge decision is opposite to the output from the code
2. Output from code matches judge decision to an extent
3. Code output and judge decision match

| Opposite result | Output somewhat matches decision | Same result | Total |
|---|---|---|---|
| 3 | 2 | 5 | 11 |

This element of the research illustrates whether the AI model produced code which was accurate to the judgement, and if it did not then this illustrates that judgement was too ambiguous, meaning that a high powering LLM could not interpret the meaning, and, as said previously, this contract should not have been drafted in such an unclear manner in the first place. Additionally, when conducting this research, for outputs that did not match the judge's decision, we also had to consider that this was due to an incorrect judge's decision – as per Marley v Rawlings which required a Supreme Court appeal for the decision to align with the code output. For code outputs that matched the legal judgement, this displays the advantages of executable contracts, as instead of a judge weighing up the meaning of ambiguous phrasing, the code symbolising the contract can be run, producing an output.

The "opposite result" category can also be further divided into two categories, which were addressed in the research table alongside the code:

1. Opposite result due to suspected LLM error
2. Opposite result due to either or due to language being to ambiguous

| LLM error | Either/Ambiguous Language | Total |
|---|---|---|
| 2 | 1 | 3 |

For this element of the research analysis, it was important to consider why the code output does not match the judge's decision, with the possibility of an incorrect judge decision being disregarded as this is incredibly unlikely. Thus, there are two major reasons why the two results do not match: LLM error or ambiguous language making it difficult to translate into C# code. For the two cases: Uber BV v Aslam and Pimlico Plumbers Ltd vs Smith, we concluded that, due to certain factors, such as Mr. Smith's obligations to Pimlico Plumbers Ltd, being disregarded, then there must have been an error by the LLM in not properly transposing the legal contract into C# code. This highlights the importance of excellent prompt engineering when dealing with an LLM, along with the mistakes that LLMs can still make. For the third case where the code output did not align with the legal judgement, (Lloyds TSB Foundation v Lloyds Banking Group), it was unclear whether the LLM was purely incorrect, or whether the LLM was incorrect due to unclear language. This was due to the output being a number which lacked context, and the code additionally required an input from the user which demonstrates that the initial contract was ambiguous in nature.

Conclusively, this analysis explored the alignment between code outputs and judge decisions and through this highlighted the positives of contracts in programming languages along with the consequences of ambiguous language in contracts. Moreover, we acknowledged that misalignments can stem from either AI errors or ambiguity in the contractual language itself. And, no matter the reason, this emphasises the need for precise prompt engineering when working with LLMs.

#### 4.1.1 Ambiguity through absence

As stated, the above section details the results table for the cases that could not be compiled into C# code, thus the table is the same as the results table in Section 3.4, however there is no C# alternative column, along with there being no design breaches in code column:

| Case Title | Summary | Case Result | Summary of Ambiguous Language | Quote of said language, with ambiguity highlighted |
|---|---|---|---|---|
| [22] Marks and Spencer plc (Appellant) v BNP Paribas Securities Services Trust Company (Jersey) Limited and another (Respondents) | This case revolves around the tenant's break clause in a lease agreement. The tenant (M&S) exercised its right to terminate the lease early. The issue relates to whether M&S is entitled to recover the apportioned rent. The resolution depends on whether a term can be implied from the lease which indicates that the apportioned rent must be refunded. | The Court of Appeal, through Lady Justice Arden, Lord Justice Jackson, and Lord Justice Fulford, concluded that the lease did not reasonably allow for such an implied term. They reasoned that from the interpretation of the lease and given the background, such a term would not naturally be read into the agreement. The judgment emphasized the importance of loyalty to the express terms of the contract and the reluctance of courts to imply terms unless absolutely necessary to achieve the parties' intentions as reflected in the contract.<br><br>Therefore, the appeal by BNP Paribas was successful, resulting in M&S not being entitled to recover the proportionate amount of rent for the period following the early termination of the lease. | The ambiguity arises from the lack of expression regarding the refund of rent paid in advance when the lease is terminated early by the tenant. The lease fails to address the apportionment of rent already paid for a period extending the termination date. This creates uncertainty surrounding the entitlement of a tenant to receive a refund, as this is not covered in the lease terms. | "Basic Rent was payable in advance on the usual quarter days... The break clause allowed the lessee to terminate the lease on... The lease, however, did not have an express term entitling the lessee to be repaid any sum by way of Basic Rent... in excess of what is due from him for periods prior to the break date." |
| [23] R v Secretary of State for the Home Department, Ex Parte Pierson | The case involves the decision concerning setting the tariff for a life sentence prisoner. The central issue was whether the Home Secretary has the authority to increase the minimum term tariff for a life sentence after it had been set in the legislation. | The House of Lords held that the Home Secretary could not retrospectively increase the minimum term. They decided that to do so would be contrary to fundamental principles of fairness and the rule of law. The reasoning was that when a court imposes a life sentence with a minimum term, it is based on the period deemed necessary for retribution and deterrence, considering the circumstances of the case. Once set by the judiciary, these tariffs should not be subject to alteration by the executive branch in a way that increases the punishment retrospectively. | The ambiguity arises from the lack of clarity in the language regarding the extent of the Home Secretary's discretion to alter a previously set tariff. Moreover, the question of whether the Home Secretary could retrospectively apply a new policy to increase the tariff was not addressed explicitly in the legislation. | "The principle of legality means that Parliament must squarely confront what it is doing and accept the political cost. Fundamental rights cannot be overridden by general or ambiguous words. This is because there is too great a risk that the full implications of such statutes may have passed unnoticed in the democratic process... In the absence of express language or necessary implication to the contrary, the courts therefore presume that even the most general words were intended to be subject to the basic rights of the individual." |
| [24] Google Inc v. Vidal-Hall | The case of Google Inc v. Vidal-Hall [2015] EWCA Civ 311 dealt with significant issues concerning data protection and privacy law. The plaintiffs, Judith Vidal-Hall and others, were internet users who alleged that Google had misused their private information by tracking their internet browsing habits without their consent through "cookies." These cookies bypassed default privacy settings on Apple's Safari web browser and allowed Google to gather information for targeted advertising. | In Google Inc v. Vidal-Hall, ambiguous language highlighted uncertainties within privacy and data protection law, particularly around what qualifies as "personal data" under the Data Protection Act 1998. This case questioned whether data collected through Google's cookies, primarily browsing habits, could be considered personal data requiring protection. Another ambiguity involved the interpretation of "damage" under Section 13 of the DPA, which traditionally meant financial loss. The Court of Appeal allowed claims for "distress" without pecuniary loss, broadening the understanding of damage under the Act. Additionally, there was uncertainty about recognizing "misuse of private information" as a tort in UK law. The case affirmed this status, enabling claimants to seek damages for privacy rights violations, emphasizing the need for privacy law to evolve with modern technology to protect individuals' rights. | The ambiguity in this case arises from Google's public statement that Safari's default settings would prevent tracking unless otherwise specified by the user, which was contradicted by the actual operation of the "safari workaround." This allowed google to bypass privacy settings and track users without consent, leading to a discrepancy between Google's stated policy and its practices. | "Further, in the premises, the Defendant's public statement (referred to at paragraph 8.3 above) about the effect of the Safari default settings upon its ability to send the DoubleClick Cookie to Safari browsers, was false." "Against this background, we cannot find any satisfactory or principled answer to the question why misuse of private information should not be categorised as a tort for the purposes of service out of the jurisdiction. Misuse of private information is a civil wrong without any equitable characteristics. |

## 4.2 Conclusion

When looking at the future of law it is difficult to imagine a change in the system. However, from a technological perspective, the integration of technological advancements in the legal sector will become ever more prevalent, with this not being limited to the use of programming languages in contracts and executable contracts. As we consider this shift, it becomes evident that this technologies could address longstanding ambiguities and inefficiencies in legal processes. For instance, programming languages could provide a precise framework for drafting contracts, thereby reducing potential litigation.

Moreover, looking beyond contracts, technological advancements propose radical innovations, such as the use of AI as a replacement to the judge. In theory, this concept would both parties agreeing on a prompt to put into a highly functioning and unbiased LLM, and then the AI model determines which party wins the case by examining the facts, much like a judge today. In principle, when discounting the algorithmic bias of the LLM, this would provide an unbiased result for the case thus enhancing judicial efficiency and impartiality.

This proposition of the move from plain language to programming language for purpose in legal contracts is not one without flaws, but reimagines the legal system by embracing evolving technology. This system promises enhances clarity, fairness and efficiency. As such, this evolution of legal contracts illustrates a future where technology and law are seamlessly intertwined with the pursuit of clarity.